\documentclass[letterpaper, 10 pt, conference]{ieeeconf}
\IEEEoverridecommandlockouts      

\usepackage{times}

\makeatletter
\let\NAT@parse\undefined
\makeatother
\usepackage[numbers,compress]{natbib}

\usepackage{multicol}
\usepackage[bookmarks=true]{hyperref}
\usepackage{amsmath} 
\usepackage{amssymb} 
\usepackage{algorithm}
\usepackage{algpseudocode}
\usepackage{graphicx}
\usepackage{array}
\usepackage{booktabs}
\usepackage{caption}
\usepackage{subcaption}
\usepackage{xcolor}
\usepackage{multirow}
\usepackage[skip=2pt,font=small]{caption}
\usepackage{makecell} 
\usepackage{paralist}
\usepackage{mathrsfs}
\usepackage{tabularx}
\usepackage{cuted}
\usepackage{colortbl}

\definecolor{Red}{rgb}{1,0,0}

\newcolumntype{P}[1]{>{\centering\arraybackslash}p{#1}}
\newcolumntype{M}[1]{>{\centering\arraybackslash}m{#1}}
\newcommand{\delete}[1]{}

\begin{document}

\title{\LARGE \bf PAIR: Proprioception-Aligned Image-Ray Representations for Viewpoint-Robust Manipulation}
\author{
Sung-Wook Lee$^{1}$, Xuhui Kang$^{1}$, Dylan Losey$^{2}$, Yen-Ling Kuo$^{1}$
  \thanks{ $^{1}$University of Virginia {\tt \{dcs3zc,xuhui,ylkuo\}@virginia.edu}}%
  \thanks{ $^{2}$Virginia Tech {\tt losey@vt.edu}}%
}

\maketitle
\setlength{\stripsep}{-15pt}   
\begin{strip}
    \vspace{4pt}           
    \centering
    \includegraphics[width=0.92\textwidth]{fig/teaser-compressed.pdf}
    \captionof{figure}{
        \textbf{Resolving visual ambiguity with camera geometry.} \textbf{(A)} Robot datasets often contain visually confounding observations, where similar robot appearances correspond to substantially different robot states. \textbf{(B)} Camera-ray information provides explicit camera-to-robot geometry to help disambiguate such observations.
    }
    \label{fig:teaser}
    \vspace{20pt}
\end{strip}

\begin{abstract}
While imitation learning has made substantial progress toward generalizable robot manipulation through scaling, viewpoint robustness remains a major bottleneck. Visuomotor policies rely heavily on visual inputs, yet images provide an inherently ambiguous signal under viewpoint change: similar scenes may appear markedly different across viewpoints, while distinct states may share certain visual overlaps. A policy must therefore infer not only the task-relevant state, but also how it is visually projected under a particular camera configuration. This entanglement of viewpoint-dependent appearance and task-relevant state can make learning inefficient and limit generalization to unseen viewpoints. Recent work has begun to address this challenge through camera-ray conditioning and representation learning using geometric cues to induce invariance. However, providing camera information as additional context does not by itself ground visual representations in how the task scene is visually projected, while geometric alignment without camera context can introduce spurious correspondences by ignoring viewpoint-dependent projection.
We propose PAIR, a contrastive learning framework that aligns joint image-ray representations using robot proprioception. PAIR encourages observations with similar robot states to remain close in representation space across viewpoints, providing a structured grounding signal for learning viewpoint-robust features. PAIR is a highly adaptable framework: it can be used to train Diffusion Policy (DP) or post-train vision-language-action (VLA) models. In simulated and real-world manipulation benchmarks with camera variation, PAIR consistently outperforms the behavior cloning baseline without ray conditioning by 12.7 \% and 26.7 \% points, respectively, demonstrating stronger generalization to unseen viewpoints. Project website: PAIRrobot.github.io.


\end{abstract}

\IEEEpeerreviewmaketitle

\begin{figure*}[t]
    \begin{center}
        {\includegraphics[width=0.905\linewidth]{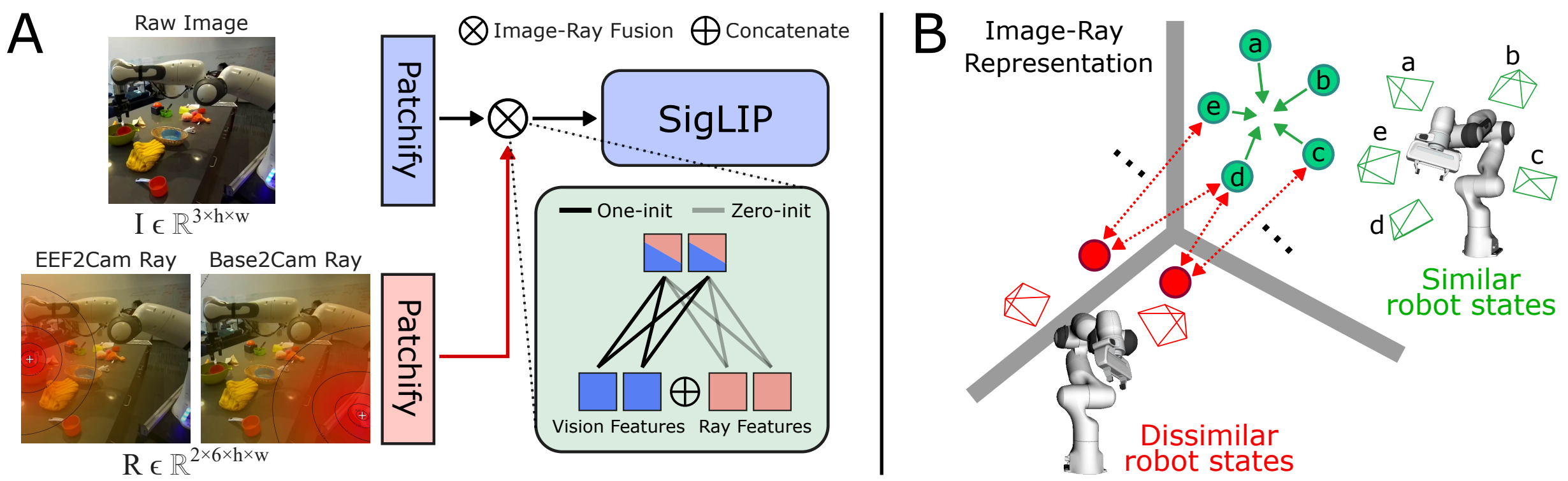} 
        }
        \vspace{2pt}
        \caption{\textbf{Overview of PAIR.} \textbf{(A)} PAIR introduces two ray representations: camera rays expressed relative to the moving end-effector frame and the robot-base frame, both encoded using Plücker coordinates. Red indicates zero Plücker moment, corresponding to rays that pass through the respective origin. The ray features are injected into the image features through an identity-initialized linear projection, which acts as a selective positional embedding for each vision token. \textbf{(B)} The joint image-ray representations are aligned according to proprioception similarity using supervised contrastive learning, which induces invariance against viewpoint variation.
        }
        \label{fig:main}%
    \end{center}
\vspace{-22pt}
\end{figure*}

\vspace{-0pt}
\section{Introduction}
Robot imitation learning has made significant progress toward general-purpose manipulation by scaling training data, model capacity, and compute~\cite{ICLR2025_88b7b2c8}. Recent visuomotor policies can learn increasingly more general manipulation behaviors directly from image observations, and can transfer across different objects, tasks, and scenes. However, robustness to viewpoint variation remains a key challenge for generalizable manipulation \cite{pang2025reviwo}.

This challenge stems from a fundamental property of visual observations: images do not directly encode the physical state of the world, but instead capture a viewpoint-dependent projection of that state. 
As the camera viewpoint changes, a fixed underlying task state can give rise to substantially different visual observations through changes in perspective and visibility. Conversely, distinct physical states may appear deceptively similar when they share similar camera-relative geometry or when the task-relevant context is only partially captured within the camera’s field of view (See \hyperref[fig]{Fig.~\ref*{fig:teaser}.A}).
This ambiguity is further exacerbated by typical camera-independent action spaces, in which visual observations vary with camera viewpoint while the corresponding action targets remain unchanged.
Consequently, to infer which actions are appropriate for the given task, a visuomotor policy must not only capture the task-relevant information in the image, but also reason about how exactly they are visually projected. 

Consider a robot approaching a mug for grasping. From one view, the gripper may appear to the left of the mug, while from another view, the same physical configuration may make the gripper appear to the right of the mug. Yet this change in image-space geometry does not imply a corresponding change in what the robot should physically do.  
The policy must therefore determine how the observed visual relationship between the robot and the mug corresponds to their underlying spatial relationship. This implicitly demands that the policy look for additional visual cues such as the robot base to establish the robot's spatial frame or fuse the vision features with the internal robot state for viewpoint-invariant action prediction. However, such visual context is not always guaranteed to be available in full, nor reliably utilized when learning from demonstrations. In addition, existing visuomotor policies typically rely on shallow, non-principled fusion between vision and robot state, where images are processed by a vision encoder and robot-state features are incorporated later through simple fusion mechanisms~\cite{chi2023diffusionpolicy,pi05_corl2025}. 

Recent studies have explored improving viewpoint invariance for visuomotor policies through camera-ray conditioning~\cite{jiang2026knowyourcamera} to use the camera state governing the visual projection as input, or use geometric signals such as proprioception to construct positive pairs for representation alignment~\cite{lee2025class,kim2026contrastive}.
However, while camera-ray conditioning provides useful viewpoint context, it does not by itself ground visual representations in the projected task state. Similarly, while alignment using physically grounded signals can encourage invariance across physically similar states, this can produce conflicting supervision by ignoring viewpoint-dependent projection without camera context. 
In light of this, we propose PAIR, a contrastive learning framework for viewpoint-robust manipulation that jointly learns image–ray representations using Plücker embeddings with supervised proprioceptive alignment. PAIR is a highly adaptable framework: it can be used to train lightweight policy classes such as diffusion policy (DP) or fine-tune pre-trained Vision-Language-Action (VLA) models. Specifically, PAIR builds on the pre-trained SigLIP-family encoders and incorporates the proposed alignment objective as an auxiliary loss alongside the standard behavior cloning objective.

We make the following contributions. First, we introduce PAIR, a general training framework that unlocks the benefit of proprioceptive alignment through joint image--ray representation learning. Second, we show that PAIR consistently improves the performance of both diffusion policies and VLA models when trained on the same demonstration data. Third, we conduct extensive evaluations across multiple simulated and real-world benchmarks under systematic viewpoint variations, demonstrating strong robustness to changes in camera viewpoint both quantitatively and qualitatively.

\section{Related Work}

\subsection{Vision Representations for Robotic Manipulation}

A dominant paradigm in visual representation learning is vision-language alignment, which has enabled VLA models with strong semantic understanding and instruction-following capabilities. Accordingly, many VLM and VLA architectures adopt pre-trained vision encoders such as CLIP~\cite{radford2021clip} and SigLIP~\cite{zhai2023siglip}, whose representations are explicitly aligned with language semantics through large-scale image-text pre-training. While this semantic alignment provides strong high-level visual understanding, such representations often lack the fine-grained spatial and geometric awareness required for robotic manipulation~\cite{ICLR2025_88b7b2c8}. 


Several methods study viewpoint robustness directly from RGB observations with data augmentation techniques including systematic data collection~\cite{xie2026multi}
or novel-view synthesis~\cite{chen2024roviaug}, while other methods improve the learning signal through cross-view supervision~\cite{seo2023multiview} and contrastive learning~\cite{lee2025class,kim2026contrastive}.
Recent efforts attempt to directly use camera extrinsic information to account for viewpoint changes. For example, \citet{zhang2026grounding} express robot actions in the camera frame to unify action representations across heterogeneous robot datasets, while \citet{jiang2026knowyourcamera} condition the policy on Plücker ray maps derived from the camera pose.
Rather than using camera geometry solely as an input signal or as a reference frame for expressing actions, PAIR grounds the image-ray representation in the underlying robot proprioceptive state using contrastive learning.

\subsection{Rays as Positional Embeddings}

Recent vision models increasingly incorporate camera rays as an explicit geometric signal for vision tasks including video generation~\cite{he2025cameractrl}, novel-view synthesis~\cite{chen2026wildrayzerselfsupervisedlargeview}, and pose estimation~\cite{zhang2024raydiffusion}. Similar lines of work have also explored ray injection as positional encoding. For example, RayPE~\cite{yin2026raype} incorporates Plücker ray geometry directly into Transformer attention for video generation. In robotics, \citet{xu2025seeing} builds on RayPE for modality-balanced action generation in VLAs, while RayViT~\cite{wang2026rayvit} injects patch-aligned Plücker ray features into a pre-trained DINOv3 ViT backbone as geometry-aware positional embeddings.
PAIR similarly introduces Plücker rays as positional embeddings, but applies this during post-training of SigLIP/SigLIP2~\cite{zhai2023siglip,tschannen2025siglip} encoders to ground the camera-ray-aware visual representations using robot proprioception.

\section{Method}
In this section, we formalize the problem in Sec. \ref{sec:problem_statement}, analyze the visual ambiguity that motivates our method in Sec. \ref{sec:visual_ambiguity}, introduce the image-ray fusion mechanism in Sec. \ref{sec:ray_alignment_theory}, and describe the training procedure in Sec. \ref{sec:supcon}.

\subsection{Problem Statement}
\label{sec:problem_statement}

We consider visuomotor policy learning from visual observations $o$, robot proprioceptive state $s$, camera rays $r$, and a task specification $g$ (e.g., language), where we augment imitation learning with a spatially grounded visual representation learning problem. Specifically, a vision encoder $f_\phi$ maps the visual observation and optionally its associated camera geometry to a latent representation
$z = f_\phi(o, r)$,
where the ray information $r$ provides a geometry-aware positional encoding for the visual tokens. We assume that camera positions change across episodes, but remained fixed within each episode.

Our goal is to learn $f_\phi$ such that is grounded in the underlying robot state $s$. In particular, observations of the same physical robot configuration from different viewpoints should induce consistent latent representations. The resulting representation is then used by a visuomotor policy
$\pi_\theta(a_{1}\mid z, s, g)$,
which predicts a sequence of future robot actions over an action horizon $H_p$.

\subsection{Visual Ambiguity of Proprioceptive Alignment}
\label{sec:visual_ambiguity}

While proprioceptive alignment of visual representations provides a potential mechanism for inducing viewpoint invariance, aligning image representations according to proprioception alone is inherently ambiguous: proprioception specifies a state in a fixed physical frame, whereas the corresponding visual observation is largely affected by the camera–robot geometry.

Let $T_{BE}$ and $T_{BC}$ denote the end-effector pose and the camera pose in the robot-base frame, respectively. The visual appearance of the robot end-effector is governed in large part by their relative transform
\begin{equation}
T_{CE}=T_{BC}^{-1}T_{BE}.
\end{equation}
Consequently, two observations $a$ and $b$ may satisfy
\begin{equation}
T_{CE}^{a}\approx T_{CE}^{b},
\qquad
T_{BE}^{a}\neq T_{BE}^{b},
\label{eq:relative_pose_aliasing}
\end{equation}
so that the robot appears similar despite occupying different physical states, as illustrated in \hyperref[fig]{Fig.~\ref*{fig:teaser}.A}. Conversely, the same physical state can appear substantially different under varied camera poses. This leads to conflicting signals for the vision encoder: visually similar observations may be pushed apart, while visually dissimilar observations are pulled together, without specifying the camera transformations responsible.

\subsection{Ray Positional Encoding}
\label{sec:ray_alignment_theory}

PAIR resolves this ambiguity by explicitly providing camera ray information, such that the encoder receives both the visual input and the camera geometry under which it was projected,
and using proprioception to align the joint image--ray representation. Specifically, we motivate camera rays as a geometry-aware positional signal for visual tokens. As observed in RayPE~\cite{yin2026raype}, Plücker ray geometry is compatible with attention mechanisms because its canonical pairwise relation is bilinear. Let the ray associated with token $i$ be
\begin{equation}
r_i=(u_i,m_i)\in\mathbb{R}^{6},
\qquad
m_i=o_i\times u_i,
\end{equation}
where $o_i$ and $u_i$ denote its origin and direction in 3D space. The Plücker reciprocal product between two rays is
\begin{equation}
B(r_i,r_j)
=
u_i^\top m_j+m_i^\top u_j,
\label{eq:plucker_reciprocal}
\end{equation}
a pairwise measure of their geometric incidence. When projected ray features enter the query and key vectors, the geometry--geometry component of an attention score takes the bilinear form
\begin{equation}
r_i^\top M r_j,
\end{equation}
where, for example, $M=\left[\begin{smallmatrix}0&I\\ I&0\end{smallmatrix}\right]$ exactly recovers Eq.~\eqref{eq:plucker_reciprocal}. Thus, attention between tokens from different views can directly capture pairwise geometric relations between their 3D rays, while naturally representing the Plücker reciprocal product.
This structure is particularly relevant in robot demonstrations, where the robot embodiment provides a persistent physical reference with consistent correspondences across viewpoints for similar proprioceptive states.
We use this compatibility to motivate exposing ray geometry to the attention layers of the vision encoder.
See Table~\ref{tab:attn_analysis} for quantitative analysis of attention activations and Figure~\ref{fig:attn_map} for qualitative visualizations of ray-conditioned attention maps.

For each visual token $x_i\in\mathbb{R}^{c}$ and each of the $n_r$ robot arms, we construct the corresponding Plücker rays and encode them using lightweight convolutional networks. Specifically, we obtain two projected camera ray embeddings: $r_i^{\mathrm{CE}_k}\in\mathbb{R}^{r}$, expressed relative to the end-effector of robot $k$, and $r_i^{\mathrm{CB}_k}\in\mathbb{R}^{r}$, expressed relative to the fixed robot base, as shown in \hyperref[fig]{Fig.~\ref*{fig:main}.A}. 
The end-effector-relative rays encode a moving target that changes with the robot's movement, whereas base-relative rays provide a persistent physical anchor across observations. These references jointly encode dynamic and static geometric structure, providing the geometric cues necessary for proprioceptive alignment across changes in visual appearance. For empirical results showing the benefits of using both ray modalities, see Table \ref{tab:ray_ablation}.

We concatenate all representations and project them back to the original
visual dimension,
\begin{equation}
\tilde{x}_i
=
W
\begin{bmatrix}
x_i \\
\rho_i^{1} \\
\vdots \\
\rho_i^{n_r}
\end{bmatrix}
+ b,
\:
\rho_i^{k}
=
\begin{bmatrix}
r_i^{\mathrm{CE}_k} \\
r_i^{\mathrm{CB}_k}
\end{bmatrix},
\:
W\in\mathbb{R}^{c\times(c+2rn_r)},
\label{eq:pair_ray_projection}
\end{equation}
where we initialize this projection as
\begin{equation}
W=
\begin{bmatrix}
I_c & 0
\end{bmatrix},
\:
b=0.
\end{equation}
The resulting image-ray patches $\tilde{x}_i$ is then added to the original position embedding of the pre-trained SigLIP model.
As a result, ray injection introduces minimal architectural change: a single identity-initialized linear layer is added to the original model architecture.

\begin{figure*}[t]

    \begin{center}
        {\includegraphics[width=0.98\linewidth]{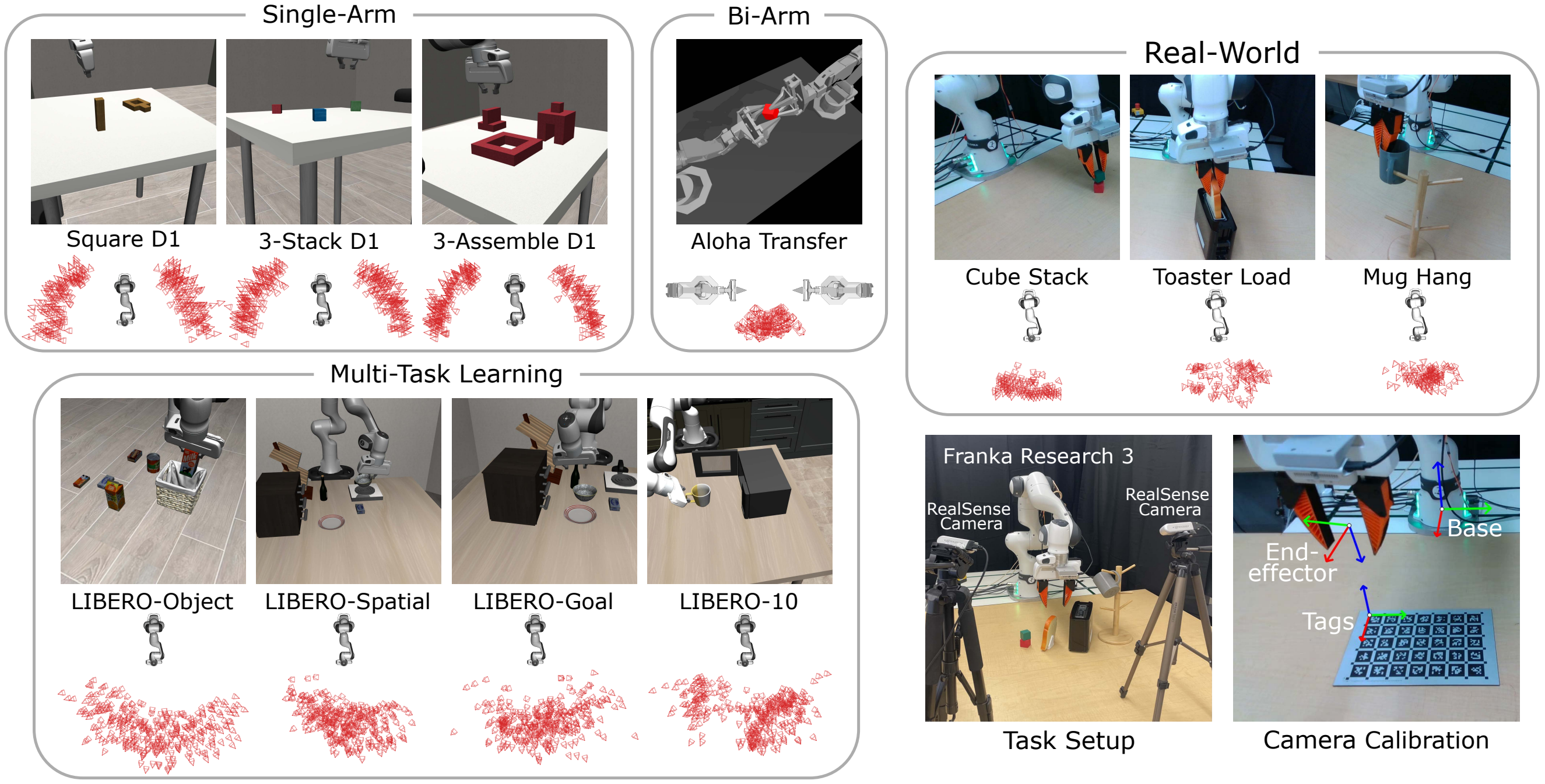}
        }
        \vspace{2pt}
        \caption{
        \textbf{Simulated benchmarks and real-world experiments.} PAIR is evaluated across both simulated and real-world manipulation tasks. In simulation, MimicGen and LIBERO use a Franka Panda arm with two side views and one front view, respectively, while AlohaGym uses a bimanual Aloha platform with two front views. In the real world, we evaluate three tabletop tasks using a Franka Research 3 (FR3) arm with two front view cameras, where the camera poses are randomized and estimated using AprilTags~\cite{olson2011apriltag} at the start of each rollout.
        }
        \label{fig:tasks}%
    \end{center}    
\vspace{-22pt}
\end{figure*}

\subsection{Supervised Contrastive Learning}
\label{sec:supcon}
For proprioception-aligned image-ray representation learning, we adopt a supervised contrastive learning~\cite{khosla2020supervised} objective based on geometric similarity between robot states. Specifically, let $\hat{z}_i$ denote the $\ell_2$-normalized latent representation of observation $i$, and let the corresponding proprioceptive state be $s_i=(p_i,q_i,g_i)$, where $p_i$, $q_i$, and $g_i$  denote the end-effector position, orientation, and gripper width, respectively.
Given a training batch of size $B$, 
we define the pairwise similarity matrix between observations $i$ and $j$  as 
\begin{equation}
S_{ij}
=
\frac{\hat{z}_i^\top \tilde{z}_j}{\tau},
\vspace{-1pt}
\end{equation}
where $\tau$ is the sharpness temperature. In the single-view setting, $\tilde{z}_j=\hat{z}_j$, while in the multi-view setting,
$\tilde{z}_j = \hat{z}^{\mathrm{cv}}_j$, which is a cross-view instance of $\hat{z}_j$. 

For spatial grounding, we define the ground-truth pairwise distance as
\vspace{-2pt}
\begin{equation}
D_{ij}
=
\|p_i-p_j\|_2
+
\lambda_q d_q(q_i,q_j)
+
|g_i-g_j|,
\label{eq:proprio_distance}
\end{equation}
where $\lambda_q$ controls the quaternion distance term
\begin{equation}
d_q(q_i,q_j)
=
2\arccos\!\left(\left|q_i^\top q_j\right|\right).
\end{equation}
For multi-robot settings, the pairwise distances between corresponding arms are averaged.
We convert the geometric distance in Eq.~\eqref{eq:proprio_distance} into a percentile-based soft positive weight. Let $\alpha_{ij}$ denote the global percentile rank of $D_{ij}$ and $\beta$ the percentile cutoff. We define
\vspace{-1pt}
\begin{equation}
w_{ij}\!=\!
\begin{cases}
1-\dfrac{\alpha_{ij}}{\beta}, & \alpha_{ij} < \beta,\\
0, & \alpha_{ij} \geq \beta,
\end{cases}
\label{eq:proprio_weight}
\vspace{-1pt}
\end{equation}
such that the closest pairs receive weight $1$ (hard positive), decreasing linearly to $0$ at the cutoff. We optimize the weighted InfoNCE objective
\begin{equation}
\begin{gathered}
\mathcal{L}_{\mathrm{PA}}\mathord{=}\mkern-1mu-\mkern-6mu
\sum_{i=1}^{B}
\mkern-4mu
\frac{\sum_{j=1}^{B} w_{ij}\log p_{ij}}
{B\cdot\sum_{j=1}^{B} w_{ij}},
\
p_{ij}\mkern2mu\mathord{=}\mkern2mu
\frac{\exp(S_{ij})}
{\sum_{k=1}^{B}\exp(S_{ik})},
\end{gathered}
\label{eq:proprio_loss}
\end{equation}
where, in the single-view setting, diagonal pairs (i=j) are excluded from the loss.
This pulls image-ray representations with similar robot states closer, while pushing apart those with dissimilar robot states, as shown in \hyperref[fig]{Fig.~\ref*{fig:main}.B}.
The final training objective combines the behavior cloning loss $\mathcal{L}_{BC}$ with the proprioception alignment loss $\mathcal{L}_{PA}$:
\vspace{-1pt}
\begin{equation}
    \mathcal{L}_{total}
    =
    \mathcal{L}_{\mathrm{BC}}
    +
    \lambda_{\mathrm{PA}}\mathcal{L}_{\mathrm{PA}},
\label{eq:pair}
\end{equation}
\vspace{-1pt}
where $\lambda_{\mathrm{PA}}$ scales the proprioception alignment objective.

\vspace{-1pt}
\section{Experiments}
In this section, we test the effectiveness of PAIR for viewpoint generalization of visuomotor policies under various settings. To understand whether the benefits arise specifically from proprioceptive alignment, we compare against temporal alignment~\cite{Sermanet2017TimeContrastiveNS}, which defines correspondence based on the temporal distance between observations. We then examine how these alignment signals interact with camera-ray conditioning and affect downstream policy learning. Specifically, we address the following research questions:

\vspace{4pt}
\begin{compactenum}
\item[\textbf{RQ1}] Does PAIR improve visual robustness across robotic embodiments (single-arm vs bi-arm), learning settings (single-task vs multi-task), and policy classes (Diffusion Policy vs VLA)?
\item[\textbf{RQ2}] Is proprioception a better alignment signal than time when learning from expert demonstrations?
\item[\textbf{RQ3}] Is ray conditioning necessary to unlock the benefits of proprioceptive alignment for visual representations?

\end{compactenum}

\vspace{-1pt}
\subsection{Benchmarks}

We evaluate PAIR on three simulated robot manipulation benchmarks spanning different embodiments, tasks, and learning settings.

\paragraph{\textbf{MimicGen}}
~\cite{mandlekar2023mimicgen} is a benchmark for imitation learning with synthetically augmented demonstrations. We use MimicGen as a standard single-task learning benchmark to study viewpoint robustness. We include the Square, Three Stack, and Three Piece Assembly tasks using the D1 distributions. Each task includes 1,000 demonstrations.

\paragraph{\textbf{AlohaGym}}
~\cite{zhao2023learning} provides a canonical bimanual manipulation task on the ALOHA platform. We evaluate PAIR on the cube transfer task, where the dataset consists of 50 human and 50 motion-planned demonstrations, totaling 100 episodes.

\paragraph{\textbf{LIBERO}}
~\cite{liu2023libero} is a manipulation benchmark comprising four task suites: Object, Goal, Spatial, and Long, which emphasize object variation, spatial reasoning, goal-conditioned behavior, and long-horizon task execution, respectively. LIBERO evaluation uses language conditioning to distinguish among tasks. Policies are trained on the full LIBERO dataset spanning 6,500 episodes in 130 tasks.

For all three benchmarks, we re-collect the original demonstrations under diverse camera viewpoints by varying the camera extrinsics while keeping the underlying task definitions unchanged. This procedure introduces controlled variation in the visual observations for a direct evaluation of visual robustness.
For single-task settings, we define the camera look-at point as the midpoint between the target objects and the robot end-effector, and sample camera distances for which the end-effector falls within the camera's field of view. 
For LIBERO, we pre-sample 50 camera poses for each task, all distinct from those used during data collection, and apply the same sampling criterion to ensure that the robot end-effector remains visible throughout each rollout. In addition, we evaluate both single-view and multi-view scenarios, where multiple viewpoints are provided for each training demonstration in the multi-view setting. A comparison of the camera settings is provided in Table~\ref{tab:env_comparison}, and the distribution of camera poses is shown in Fig.~\ref{fig:tasks}.

\begin{table}[b!]
\centering
\small
\vspace{-22pt}
\caption{Hyperparameters for Diffusion Policy and SmolVLA.}
\vspace{2pt}
\begin{minipage}[t]{0.46\columnwidth}
\vspace{0pt}
\centering
\begin{tabular}{@{}l@{\hspace{1em}}c@{}}
\toprule
\textbf{Name} & \textbf{Value} \\
\midrule
\textbf{Training} & \\
Train steps & 200K \\
Learning Rate & 1$e^{-4}$ \\
BC batch size & 64 \\
PAIR batch size & 192 \\
$\lambda_q$ & 0.05 \\
$\lambda_{\mathrm{PA}}$ & 0.02 \\
$\beta$ & 0.025 \\
$\tau$ & 0.05 \\
Pre-trained MAP & True \\
$T_o$ & 1 \\
$T_p$ & 32 \\
\bottomrule
\end{tabular}

\par\medskip
\vspace{-3pt}
\textbf{(a) Diffusion Policy.}
\end{minipage}
\hfill
\begin{minipage}[t]{0.46\columnwidth}
\vspace{0pt}
\centering
\begin{tabular}{@{}l@{\hspace{1em}}c@{}}
\toprule
\textbf{Name} & \textbf{Value} \\
\midrule
\textbf{Training} & \\
Train steps & 160K \\
Learning Rate & 1$e^{-4}$ \\
BC batch size & 64 \\
PAIR batch size & 192 \\
$\lambda_q$ & 0.05 \\
$\lambda_{\mathrm{PA}}$ & 0.05 \\
$\beta$ & 0.025 \\
$\tau$ & 0.05 \\
Pre-trained MAP & False \\
$T_o$ & 1 \\
$T_p$ & 50 \\
\bottomrule
\end{tabular}
\par\medskip
\vspace{-3pt}
\textbf{(b) SmolVLA.}
\end{minipage}
\label{tab:hyperparam}
\end{table}

\subsection{Policy Architectures}

To test whether the benefits of PAIR extend beyond a particular policy type, we evaluate it with two distinct policy classes: vision-based Diffusion Policies (DP) for single task learning and pre-trained Vision-Language-Action models for multi-task learning.

\paragraph{\textbf{Diffusion Policy (DP)}}
We use a U-Net Diffusion Policy~\cite{chi2023diffusionpolicy} with a pre-trained SigLIP2~\cite{tschannen2025siglip} visual encoder for single-task evaluation on MimicGen and AlohaGym. We apply PAIR ray conditioning to the visual tokens as described in Sec.~\ref{sec:ray_alignment_theory}. We retain SigLIP2's multi-head attention pooling (MAP) projection both for proprioceptive alignment and for aggregating the token representations into a global visual feature passed to the policy head.

\begin{table}[b]
\centering
\vspace{-8pt}
\caption{Comparison of environments, including robot and camera counts, viewpoint, multi-task setting, and task horizon. MimicGen uses separate task horizons for three tasks.}
\vspace{3pt}

\label{tab:env_comparison}
\small
\setlength{\tabcolsep}{3pt}
\renewcommand{\arraystretch}{1.1}
\resizebox{0.95\linewidth}{!}{%
\begin{tabular}{lccccc}
\hline
\textbf{Env.} &
\textbf{\# Robots} &
\textbf{\# Cams} &
\textbf{View} &
\textbf{Multi-task} &
\textbf{Horizon} \\
\hline
\noalign{\vskip 2pt}
MimicGen   & 1 & 2 & Side & No  & 800/800/1000 \\
ALOHAGym   & 2 & 2 & Front    & No  & 800 \\
LIBERO     & 1 & 1 & Front    & Yes & 600 \\
\midrule
\midrule
Real-World & 1 & 2 & Front    & Yes & 1000 \\
\noalign{\vskip 1pt}\hline
\end{tabular}}
\vspace{2pt}
\end{table}

\begin{table*}[t]
\centering
\small
\setlength{\tabcolsep}{5pt}
\vspace{-2pt}
\caption{Success rate (\%) of \textbf{Diffusion Policy (DP)} on three MimicGen tasks and ALOHA Transfer across
3 seeds. For each seed, the maximum success rates from 5 checkpoints is reported.
}
\label{tab:single-task-results}
\resizebox{0.94\linewidth}{!}{%
\begin{tabular}{llcccccccc}
\toprule
& & \multicolumn{2}{c}{\textbf{Square D1}} & \multicolumn{2}{c}{\textbf{3-Stack D1}} & \multicolumn{2}{c}{\textbf{3-Pc Assembly D1}} & \multicolumn{2}{c}{\textbf{ALOHA Transfer}} \\
\cmidrule(lr){3-4}\cmidrule(lr){5-6}\cmidrule(lr){7-8}\cmidrule(lr){9-10}
\textbf{Views} & \textbf{Auxillary Objective} & \textit{w/o Ray} & \textit{w/ Ray} & \textit{w/o Ray} & \textit{w/ Ray} & \textit{w/o Ray} & \textit{w/ Ray} & \textit{w/o Ray} & \textit{w/ Ray} \\
\midrule
\multirow{3}{*}{Single}
& None \hfill (BC)
& 11.7 $\pm$ 1.2
& 13.4 $\pm$ 2.1
& 20.1 $\pm$ 1.1
& 34.1 $\pm$ 0.9
& 10.2 $\pm$ 1.3
& 18.1 $\pm$ 2.1
& 25.3 $\pm$ 1.6
& 35.2 $\pm$ 2.9 \\

& Temporal alignment \hfill (TCN)
& 11.7 $\pm$ 1.5
& 16.3 $\pm$ 4.5
& 19.0 $\pm$ 2.0
& 39.0 $\pm$ 2.6
& 12.7 $\pm$ 2.1
& 18.3 $\pm$ 2.5
& 27.3 $\pm$ 1.5
& 35.3 $\pm$ 3.5 \\

& Proprio alignment \hfill (PAIR)
& 20.7 $\pm$ 2.8
& \textbf{24.4} $\pm$ 2.9
& 36.3 $\pm$ 2.5
& \textbf{45.1} $\pm$ 3.7
& 20.1 $\pm$ 3.5
& \textbf{29.1} $\pm$ 1.8
& 30.0 $\pm$ 0.7
& \textbf{42.3} $\pm$ 2.1 \\

\cmidrule(lr){1-10}
\multirow{3}{*}{Multi}
& None \hfill (BC)
& 26.9 $\pm$ 4.6
& 24.1 $\pm$ 1.1
& 44.1 $\pm$ 1.8
& 49.6 $\pm$ 2.6
& 17.3 $\pm$ 1.5
& 27.1 $\pm$ 1.8
& 55.7 $\pm$ 2.1
& 66.3 $\pm$ 2.9 \\

& Temporal alignment \hfill (TCN)
& 28.0 $\pm$ 2.0
& 32.3 $\pm$ 2.5
& 47.3 $\pm$ 2.1
& 50.0 $\pm$ 3.0
& 17.7 $\pm$ 2.1
& 25.3 $\pm$ 2.1
& 58.7 $\pm$ 3.2
& 70.0 $\pm$ 5.3 \\

& Proprio alignment \hfill (PAIR)
& 34.7 $\pm$ 3.0
& \textbf{38.5} $\pm$ 3.4
& 48.9 $\pm$ 3.7
& \textbf{53.9} $\pm$ 2.8
& 23.2 $\pm$ 3.4
& \textbf{30.5} $\pm$ 3.4
& 51.0 $\pm$ 2.0
& \textbf{74.0} $\pm$ 2.6 \\
\bottomrule
\end{tabular}
}
\vspace{-2pt}
\end{table*}

\begin{table*}[h]
\centering
\small
\setlength{\tabcolsep}{4pt}
\caption{Success rate (\%) of \textbf{SmolVLA} on the four LIBERO task suites across 3 seeds. For each seed, the maximum success rates from 5 checkpoints is reported.}
\label{tab:libero-results}
\resizebox{\linewidth}{!}{%
\begin{tabular}{lcccccccc@{\hspace{10pt}}cc}
\toprule

& \multicolumn{2}{c}{\textbf{Object}}
& \multicolumn{2}{c}{\textbf{Goal}}
& \multicolumn{2}{c}{\textbf{Spatial}}
& \multicolumn{2}{c@{\hspace{10pt}}}{\textbf{Long}}
& \multicolumn{2}{c}{\textbf{Average}} \\

\cmidrule(lr){2-3}
\cmidrule(lr){4-5}
\cmidrule(lr){6-7}
\cmidrule(lr{6pt}){8-9}
\cmidrule(l{6pt}r){10-11}

\textbf{Auxiliary Objective}
& \textit{w/o Ray} & \textit{w/ Ray}
& \textit{w/o Ray} & \textit{w/ Ray}
& \textit{w/o Ray} & \textit{w/ Ray}
& \textit{w/o Ray} & \textit{w/ Ray}
& \textit{w/o Ray} & \textit{w/ Ray} \\

\midrule
None \hfill (BC)                  & 94.6 $\pm$ 0.3 & 92.9 $\pm$ 0.6 & 90.4 $\pm$ 0.6 & 89.2 $\pm$ 0.5 & 82.8 $\pm$ 0.9 & 84.3 $\pm$ 1.1 & 57.9 $\pm$ 1.1 & 66.9 $\pm$ 0.9 & \textit{81.4} & \textit{83.3} \\
Time-contrastive \hfill (TCN)     & 93.3 $\pm$ 0.4 & 94.1 $\pm$ 0.3 & 88.5 $\pm$ 0.6 & 87.3 $\pm$ 0.9 & 79.5 $\pm$ 0.5 & 84.3 $\pm$ 1.1 & 61.1 $\pm$ 1.1 & 70.4 $\pm$ 0.6 & \textit{80.6} & \textit{84.0} \\
Proprio-contrastive \hfill (PAIR) & 92.8 $\pm$ 0.5 & \textbf{95.7} $\pm$ 0.3 & 84.7 $\pm$ 0.5 & \textbf{90.8} $\pm$ 0.6 & 84.5 $\pm$ 0.6 & \textbf{86.8} $\pm$ 0.5 & 73.4 $\pm$ 3.1 & \textbf{82.1} $\pm$ 0.7 & \textit{83.9} & \textit{\textbf{88.9}} \\
\bottomrule
\end{tabular}
}
\vspace{-15pt}
\end{table*}

\paragraph{\textbf{\boldmath SmolVLA}}
For multi-task learning, we fine-tune SmolVLA~\cite{shukor2025smolvla}, a pre-trained vision-language-action model. Since its SigLIP vision encoder does not retain the original MAP projection head, we initialize it from scratch. SmolVLA is trained as a language-conditioned multi-task policy and evaluated across all four LIBERO task suites.

\subsection{Training Objectives and View Settings}
All methods share the same backbone, optimizer, and schedule, and differ only
in the auxiliary objective added to the BC loss. The policy always consumes a
single camera view; single-view and multi-view settings differ only in how the
remaining views are used during training. Critically, proprioception features used to align the vision features in PAIR is provided as extra input modality for all cases.

\paragraph{\textbf{BC}}
Standard behavior cloning without auxiliary loss. In the multi-view setting, the input view is randomly pooled from the candidate views.

\paragraph{\textbf{TCN}}
The TCN~\cite{Sermanet2017TimeContrastiveNS} objective as the auxiliary loss. We follow the original implementation for both the single-view and multi-view settings.

\paragraph{\textbf{PAIR}} 
Proprioceptive alignment loss in Eq.~\eqref{eq:proprio_loss} as the auxiliary loss. 
In the multi-view setting, the target view is sampled from the set of candidate camera views to provide cross-view learning signals.

\subsection{Training/Evaluation Protocol}
Diffusion Policies are trained for 200k steps, with five checkpoints saved at 10k-step intervals from 160k to 200k steps. SmolVLA models are trained for 160k steps, with five checkpoints saved at 10k-step intervals from 120k to 160k steps. For image augmentation, we randomly crop $256 \times 256$ images and their corresponding Plücker embeddings to $224 \times 224$ during training and apply a center crop during inference. 

For evaluation, single-task policies are rolled out 200 times for each model checkpoint. For LIBERO, each task is evaluated over 50 rollouts; with 10 tasks per task suite, this totals 500 rollouts per suite. The action horizon for open-loop rollout $T_a$ is 12 for all cases except AlohaGym, which uses 16. Table~\ref{tab:hyperparam} summarizes the training and evaluation hyperparameters for both policy classes.

\subsection{Simulation Results}

Tables~\ref{tab:single-task-results} and~\ref{tab:libero-results} summarize the performance of PAIR across single-task and multi-task learning settings, respectively. 
In single-task learning, PAIR with ray conditioning improves the maximum success rate on average by 16.2\%, 9.1\%, and 9.5\% points over BC without ray conditioning, ray-conditioned BC, and PAIR without ray conditioning, respectively. On LIBERO, PAIR with ray conditioning achieves the highest average success rate of 88.9\%, outperforming ray-conditioned BC by 5.6\% points. Compared with TCN, PAIR further improves the maximum success rate by 4.9\% and 3.3\% points with and without ray conditioning, respectively.

\subsection{Real-World Experiments}
\begin{figure}[h!]
\vspace{-4pt}
    \begin{center}
        {\includegraphics[width=\linewidth]{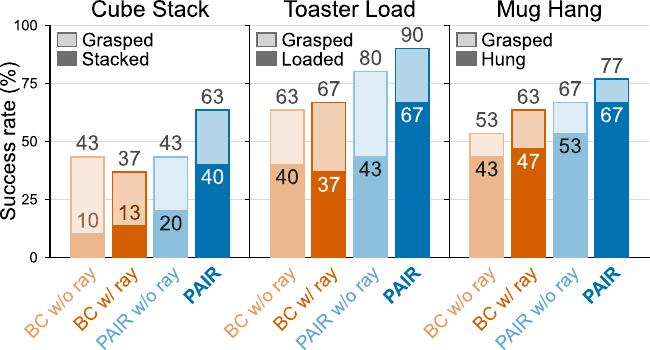}
        }
        \vspace{-10pt}
        \caption{
        \textbf{Success rates (\%) of BC (red) and PAIR (blue), with and without ray conditioning, in three real-world tasks.} Both subtask completion rates and final task success rates are reported. 
        }
        \label{fig:rw_results}%
    \end{center}    
\vspace{-20pt}
\end{figure}

We design three real-world tasks using a Franka Research 3 robot 
and two RealSense D435 cameras capturing images at 30 FPS, as illustrated in Fig. \ref{fig:tasks}. For each of the three tasks, 100 demonstrations are collected using VR-controlled teleoperation. For each rollout, cameras are re-positioned, and their poses are estimated using AprilTags~\cite{olson2011apriltag}. Similar to LIBERO, we fine-tune the pre-trained SmolVLA on all three tasks jointly with language conditioning. We compare BC and PAIR with and without ray conditioning, using 30 rollouts per task in each setting. 
To ensure that evaluation is conducted from genuinely unseen viewpoints, after each camera calibration, we resample a camera pose if both its camera position and ground-plane look-at point are within 1 cm of those of any viewpoint sampled during data collection.
As shown in Fig.~\ref{fig:rw_results}, PAIR with ray conditioning achieves the highest success rates across all three tasks, averaging 76.7\% and 57.8\% points on subtask and final task completion, respectively. This improves over BC without ray conditioning by 23.3\% and 26.7\% points, ray-conditioned BC by 21.1\% and 25.6\% points, and PAIR without ray conditioning by 13.3\% and 18.9\% points, respectively.

\subsection{Ablation Studies and Analysis}

\subsubsection{\textbf{Importance of Ray Conditioning for Proprioceptive Alignment}}
We ablate the two camera-ray references used in PAIR to assess their individual contributions to proprioceptive alignment. Table~\ref{tab:ray_ablation} shows that removing either the end-effector-relative or base-relative ray modalities consistently lead to performance drop, which suggests that the both ray modalities provide complementary geometric information.

\begin{table}[b!]
\centering
\vspace{-12pt}
\caption{Ablation of end-effector-relative ($r^{\mathrm{CE}}$) and robot-base-relative ($r^{\mathrm{CB}}$) camera rays in PAIR. Average success rates (\%) are reported, with the performance drop shown in parentheses.}

\label{tab:ray_ablation}
\small
\setlength{\tabcolsep}{6pt}
\resizebox{\linewidth}{!}{%
\begin{tabular}{lccc}
\hline
\noalign{\vskip 2pt}

\textbf{Task (View)}
& \textbf{PAIR}
& \textbf{PAIR w/o $\boldsymbol{r}^{\boldsymbol{\mathrm{CB}}}$}
& \textbf{PAIR w/o $\boldsymbol{r}^{\boldsymbol{\mathrm{CE}}}$} \\

\noalign{\vskip 1pt}
\hline
\noalign{\vskip 2pt}

Square D1 (Multi) & 38.5 & 28.8 {\scriptsize\textcolor{red!70!black}{(-9.7)}} & 35.3 {\scriptsize\textcolor{red!70!black}{(-3.2)}} \\
Aloha Transfer (Multi) & 74.0 & 55.2 {\scriptsize\textcolor{red!70!black}{(-18.8)}} & 67.0 {\scriptsize\textcolor{red!70!black}{(-7.0)}} \\
LIBERO & 88.9 & 87.2 {\scriptsize\textcolor{red!70!black}{(-1.7)}} & 86.8 {\scriptsize\textcolor{red!70!black}{(-2.1)}} \\

\hline
\end{tabular}
}
\end{table}

\subsubsection{\textbf{Effect of $\boldsymbol{\lambda_q}$ and $\boldsymbol{\lambda_{PA}}$}}
\begin{figure}[h!]
\vspace{-6pt}
    \begin{center}
        {\includegraphics[width=\linewidth]{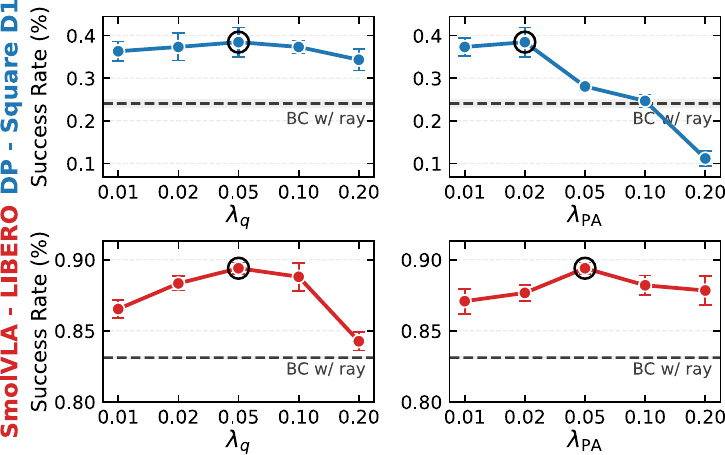}
        }
        \vspace{-12pt}
        \caption{
        \textbf{Policy performance with varied $\lambda_{q}$ and $\lambda_{PA}$ values for Diffusion Policy trained on Square D1 (blue) and SmolVLA trained on LIBERO (red) with the PAIR objective. The baselines report the performance of BC with ray conditioning.}
        }
        \label{fig:abl}%
    \end{center}    
\vspace{-10pt}
\end{figure}

In Fig.~\ref{fig:abl}, we examine the effects of the quaternion-distance coefficient $\lambda_q$ and the alignment-loss coefficient $\lambda_{\mathrm{PA}}$ on policy performance of DP trained on the multi-view Square D1 task and SmolVLA trained on LIBERO. PAIR improves performance across different choices of $\lambda_{q}$, with the best performances achieved when $\lambda_{q} = 0.05$ for both cases. For the alignment loss coefficient $\lambda_{PA}$, 0.02 and 0.05 yield the best performance for DP and SmolVLA, respectively. Notably, DP is more sensitive to larger $\lambda_{PA}$, where the alignment loss dominates the imitation learning loss.

\subsubsection{\textbf{Attention Activation Analysis}}
\label{sec:attn_analysis}
\begin{figure}[t]
    \begin{center}
    \vspace{5pt}
        {\includegraphics[width=\linewidth]{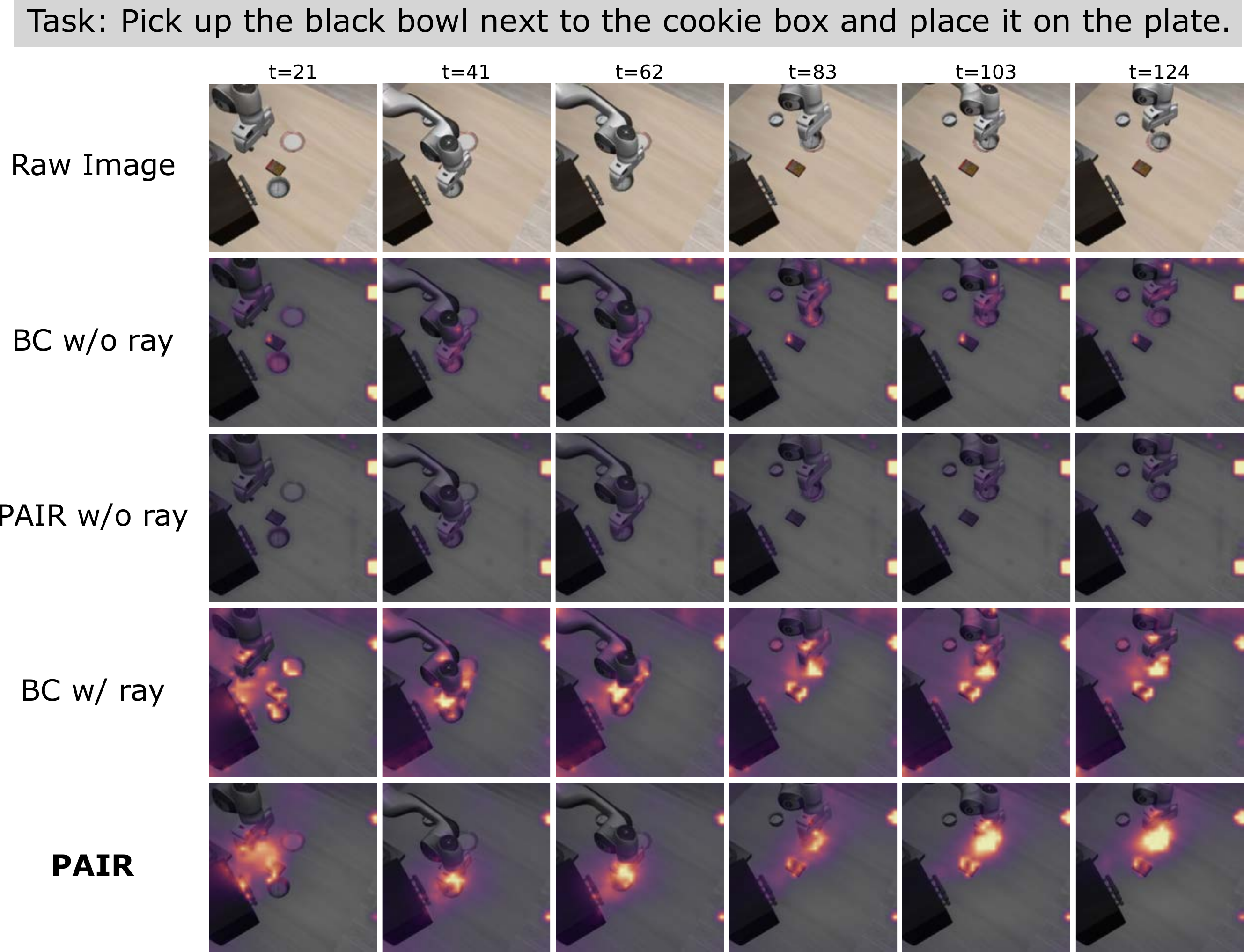}
        }
        \vspace{-6pt}
        \caption{
        \textbf{Visualization of the rollout attention maps of SmolVLA models trained on the LIBERO dataset.}
        The average attention activations are overlaid on the corresponding input images. To ensure the consistency of the rollout, we sample a single trajectory from the training dataset. Without ray input, attention is concentrated on spatially irrelevant regions, even with proprioceptive alignment. In contrast, ray-conditioning leads to stronger attention to the end effector, with proprioceptive alignment further concentrating activation around it and its nearby workspace.}
        \label{fig:attn_map}%
    \end{center}    
    \vspace{-6pt}
\end{figure}

To assess whether PAIR improves attention to the robot end-effector and nearby task region, we measure the attention assigned to the visual token containing the ground-truth tool center point across SigLIP layers in Table~\ref{tab:attn_analysis}. Ray conditioning consistently increases attention to the tool tip, with PAIR yielding the highest activation—2.51$\times$ that of BC without ray. The corresponding attention maps are visualized in Fig.~\ref{fig:attn_map}.

\begin{table}[h]
\centering
\vspace{-2pt}
\caption{Normalized average attention assigned to the visual token containing the ground-truth tool center point. The raw attention score is normalized by the expected attention under a uniform distribution, where a value of 1 denotes the expected value.}
\label{tab:attn_analysis}
\small
\setlength{\tabcolsep}{5pt}
\resizebox{0.8\linewidth}{!}{%
\begin{tabular}{lcccc}
\hline
\noalign{\vskip 2pt}
& \multicolumn{2}{c}{\textbf{BC}} 
& \multicolumn{2}{c}{\textbf{PAIR}} \\
\noalign{\vskip 1pt}
\cline{2-3} \cline{4-5}
\noalign{\vskip 1pt}
\textbf{Task} 
& w/o Ray 
& w/ Ray 
& w/o Ray
& w/ Ray \\
\noalign{\vskip 1pt}
\hline
\noalign{\vskip 2pt}
LIBERO     & 1.28 & 2.91 & 1.64 & \textbf{3.52} \\
Real-World & 1.17 & 2.16 & 2.25 & \textbf{2.67} \\
\hline
\end{tabular}
}
\vspace{-6pt}
\end{table}

\subsubsection{\textbf{Computational Cost}}
\label{sec:compute_analysis}
Table~\ref{tab:compute_analysis} compares the training cost of BC, TCN, and PAIR. While both TCN and PAIR introduce additional overhead relative to BC due to their auxiliary alignment objectives, PAIR is substantially more efficient than TCN, requiring 58.8\% fewer FLOPs and 60\% fewer decoded frames. In addition, ray input incurs only a small computational overhead, increasing FLOPs by 3.0\% for BC and 3.8\% for TCN and PAIR.

\begin{table}[h]
\centering
\caption{Computational and dataloading cost of training SmolVLA with BC, TCN, and PAIR. FLOPs are reported per forward pass for models without and with ray conditioning, respectively. Decoded Frames denotes the number of camera frames decoded during dataloading for each training batch. Training settings are provided in Table~\ref*{tab:hyperparam}.B.
}
\vspace{2pt}
\label{tab:compute_analysis}
\small
\setlength{\tabcolsep}{6pt}
\resizebox{0.93\linewidth}{!}{%
\begin{tabular}{lccc}
\hline
\noalign{\vskip 2pt}

\textbf{Metric}
& \textbf{BC}
& \textbf{TCN}
& \textbf{PAIR} \\

\noalign{\vskip 1pt}
\hline
\noalign{\vskip 2pt}

FLOPs ($\times 10^{13}$, $\downarrow$)
& 1.68 / 1.73 & 14.1 / 14.7 & 5.83 / 6.05 \\

Decoded Frames ($\downarrow$)
& 64 & 640 & 256 \\

\hline
\end{tabular}
}
\vspace{-14pt}
\end{table}

\section{Discussion} 
\label{sec:discussion}
The empirical results show that proprioceptive alignment is not uniformly beneficial on its own, with mixed gains in the absence of ray conditioning. In contrast, when combined with rays, PAIR consistently outperforms the baseline methods across all evaluated settings, which indicates that camera-ray conditioning is a computationally cheap lever for unlocking the benefits of proprioceptive alignment.

In addition, the alignment with respect to proprioceptive signal consistently outperforms the temporal signal, while ray-conditioning benefits both methods. Notably, these gains hold across different robotic embodiments, single- and multi-task learning settings, and both Diffusion Policy and VLA policy classes, indicating that the benefits of PAIR generalize across diverse robotic learning settings.

In the multi-view learning setting, PAIR improves upon its single-view counterpart by leveraging observations from multiple viewpoints. The improvement reflects PAIR’s ability to exploit cross-view signals from observations of the same task scene for proprioceptive alignment. By effectively extracting these shared cross-view signals, PAIR shows the potential to scale to more diverse viewpoint settings and support cross-view representation learning.

\section{Conclusion}
\label{sec:conclusion}
In this work, we presented PAIR, a framework for aligning image-ray representations with robot proprioception to improve viewpoint robustness.
While PAIR significantly improves generalization to unseen camera poses, 
there are several limitations. First, existing robotic datasets often contain demonstrations collected from a limited or skewed distribution of camera viewpoints, and the effectiveness of PAIR on such datasets may depend on the degree of viewpoint diversity present in the data. Second, collecting data under systematically diverse viewpoints with accurate camera calibration requires additional human effort and increases data-collection time. Finally, as shown in Table \ref{tab:compute_analysis}, the proprioceptive alignment objective incurs extra computational overhead during training. Future directions include building large-scale real-world datasets with systematically controlled viewpoint diversity, leveraging camera pose estimation for scalable image–ray pre-training on existing robotic datasets, and extending PAIR to cross-embodiment learning.

\bibliographystyle{IEEEtranN}
\bibliography{references}

\clearpage
\end{document}